\documentclass{article}

\usepackage{bm}
\usepackage{amssymb}
\usepackage{amsmath}
\usepackage{amsthm}
\usepackage{bbding}
\usepackage{pifont}
\usepackage{hyperref}
\usepackage{url}
\usepackage{graphicx}
\usepackage{booktabs}
\usepackage{xcolor}
\usepackage{multirow}
\usepackage{bbm}

\usepackage{wrapfig}
\usepackage{subfig}
\usepackage[super]{nth}
\usepackage{float}
\usepackage{algorithm}
\usepackage{algorithmicx}
\usepackage[noend]{algpseudocode}
\usepackage{cleveref}
\usepackage{color, colortbl}
\usepackage{authblk}
\usepackage[bb=boondox]{mathalfa}

\usepackage[misc]{ifsym} 

\crefname{section}{Sec.}{Secs.}
\Crefname{section}{Section}{Sections}
\Crefname{table}{Table}{Tables}
\crefname{table}{Tab.}{Tabs.}
\crefname{algorithm}{Alg.}{Algs.}
\crefname{figure}{Fig.}{Figs.}
\crefname{equation}{Eqn.}{Eqns.}

\PassOptionsToPackage{numbers, compress}{natbib}
\usepackage[preprint]{neurips_2024}

\usepackage[utf8]{inputenc} % allow utf-8 input
\usepackage[T1]{fontenc}    % use 8-bit T1 fonts
\usepackage{hyperref}       % hyperlinks
\usepackage{url}            % simple URL typesetting
\usepackage{booktabs}       % professional-quality tables
\usepackage{amsfonts}       % blackboard math symbols
\usepackage{nicefrac}       % compact symbols for 1/2, etc.
\usepackage{microtype}      % microtypography
\usepackage{xcolor}         % colors

\hypersetup{
    colorlinks=true,
    linkcolor=red,
    citecolor=cyan,
    filecolor=magenta,      
    }

\title{ZeroSmooth: Training-free Diffuser Adaptation for High Frame Rate Video Generation}

\author[1, 2]{Shaoshu Yang}
\author[3,\Letter]{Yong Zhang}
\author[3]{Xiaodong Cun}
\author[3]{Ying Shan}
\author[1, 2, \Letter]{Ran He}

\affil[1]{School of Artificial Intelligence, University of Chinese Academy of Sciences}
\affil[2]{New Laboratory of Pattern Recognition (NLPR), Institute of Automation, Chinese Academy of Sciences}
\affil[3]{Tencent AI Lab}
\affil[]{\url{https://ssyang2020.github.io/zerosmooth.github.io}}
\begin{document}

\maketitle

\let\thefootnote\relax\footnotetext{
${^{~\textrm{\Letter}}}$Corresponding author}

\begin{center}
  \captionsetup{type=figure}
  \includegraphics[width=1.00\textwidth]{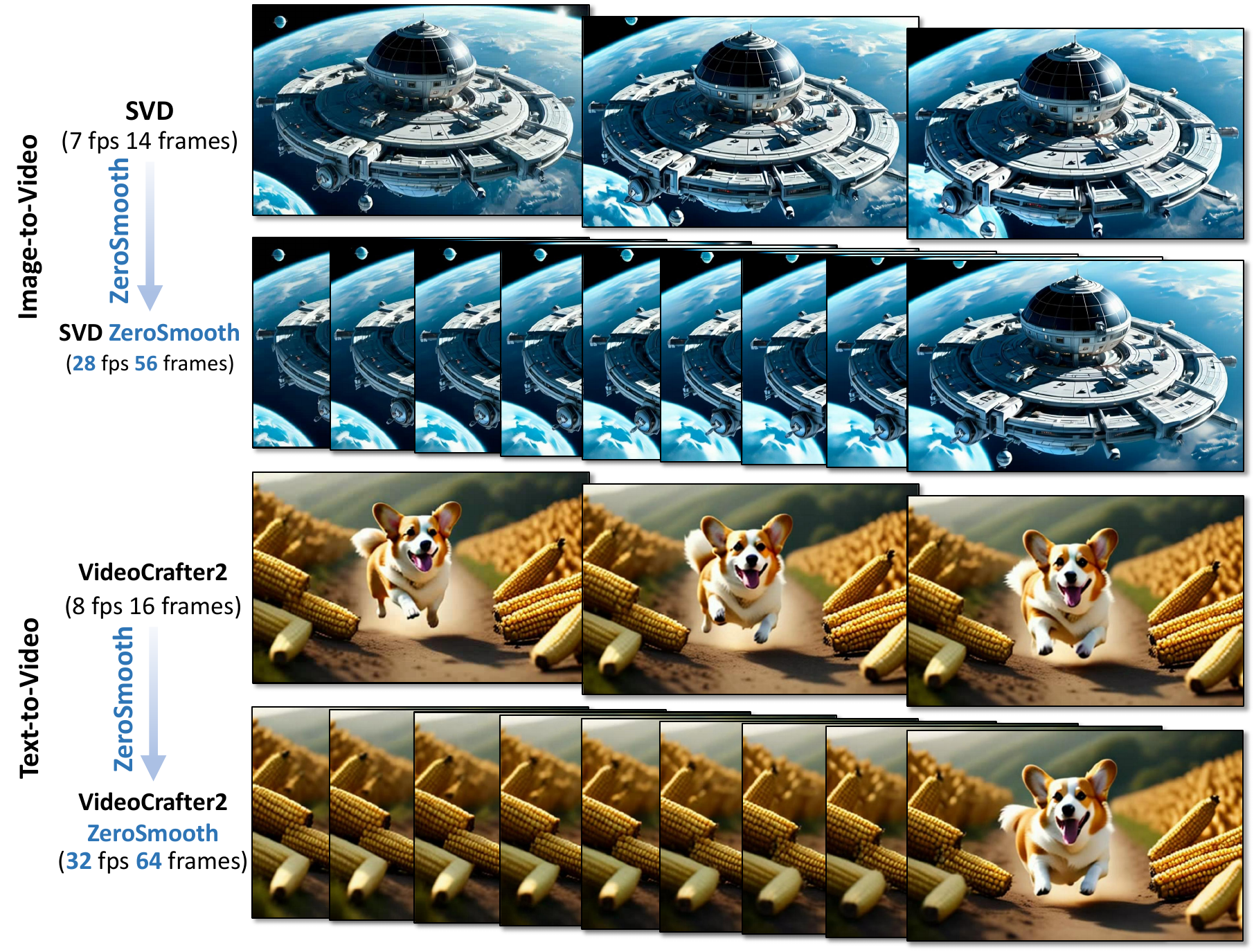}
  \captionof{figure}{
    Our method enables pretrained video diffusion models for high frame rate (4$\times$ more than during training) generation without extra training data and parameter updates.
  }
  \label{fig:teaser}
\end{center}

% Training Free Interpolation for Generative Video Models
%  - camera shooted with high-frame rate 
%  - generative video models' frame rate is not enough.
% Conventional methods: 
%   - interpolation existing videos 
%   - requires training 
%   - backward warping, forward splatting, predict conv kernels 
% LDM based methods 
%   - video inpainting 
% Our method 
%   - training free 
%   - generalizable 
%   - attention 

\begin{abstract}
Video generation has made remarkable progress in recent years, especially since the advent of the video diffusion models.
Many video generation models can produce plausible synthetic videos, \textit{e.g.,} Stable Video Diffusion (SVD). 
However, most video models can only generate low frame rate videos due to the limited GPU memory as well as the difficulty of modeling a large set of frames. 
The training videos are always uniformly sampled at a specified interval for temporal compression. 
Previous methods promote the frame rate by either training a video interpolation model in pixel space as a postprocessing stage or training an interpolation model in latent space for a specific base video model.   
In this paper, we propose a training-free video interpolation method for generative video diffusion models, which is generalizable to different models in a plug-and-play manner.
We investigate the non-linearity in the feature space of video diffusion models and transform a video model into a self-cascaded video diffusion model with incorporating the designed hidden state correction modules.  
The self-cascaded architecture and the correction module are proposed to retain the temporal consistency between key frames and the interpolated frames. 
Extensive evaluations are preformed on multiple popular video models to demonstrate the effectiveness of the propose method, especially that our training-free method is even comparable to trained interpolation models supported by huge compute resources and large-scale datasets.

\end{abstract}

% \section{Introduction}
% Introduce video diffusion models, interpolation models...

% Our contributions are summarized as follows:
% \begin{itemize}
%     \item We 
%     \item We propose ZeroSmooth, 
%     \item We deploy our method on a variety of prevelent video diffusers including VideoCrafter2, LaVie and StableVideoDiffusion, showcasing the ...
% \end{itemize}

% \section{Related Works}

% \section{Preliminary}
% \subsection{Video diffusion models}

% \subsection{Visual restoration with diffusion models}

% \section{Method}
% \subsection{Self-cascascaded video diffusion models}
% \subsection{Back-projection method for transformers}
% \paragraph{Back-projection in spatial transformers}
% \paragraph{Back-projection in temporal transformers}
% \subsection{Restore input integrity}

% \section{Experiment}
% \subsection{Comparison to baselines}
% \subsection{Comparison to cascaded video diffusers}
% \subsection{Comparison to video interpolation methods}

% \section{Conclusion}

\section{Introduction}
Video generation has undergone rapid evolution recently, benefiting from the development of diffusion models~\cite{ddpm, score-sde}.
There are numerous popular video diffusion models~\cite{video-dm} that can generate plausible synthetic videos. 
They are trained on large-scale datasets, including videos and images, using extensive compute resources. 
Several models are trained in the pixel space, \textit{e.g.,} Imagen Video~\cite{imagen-video} and Make-A-Video~\cite{make-a-video}.
The majority are trained in the latent space, \textit{e.g.,} Stable Video Diffusion (SVD)~\cite{stable-video-diffusion}, Align-your-latent~\cite{video-ldm}, VideoCrafter~\cite{chen2024videocrafter2}, LaVie~\cite{wang2023lavie}, ModelScope~\cite{modelscope}, etc.

Although those video models can produce realistic videos, most of them cannot generate high frame rate videos with smooth transitions.
One reason is that most video generation models use uniformly sampled videos as training data to avoid exceeding the GPU memory limit; \textit{i.e.,} models are trained to generate key frames.
The other reason is that capturing the distribution of a long video is challenging.
In this work, we focus on improving the frame rate of those generative video models.

Conventional video frame interpolation methods can be grouped into two categories: \textit{i.e.,} flow-based~\cite{flow1, flow2, reda2022film, flow4} and kernel-based methods~\cite{lu2022kernel1, zhang2023kernel2}.
The former follows a two-stage paradigm that first estimates flows between neighboring frames and then applies forward or backward warping to the pixels or latent features for intermediate frame generation.
The performance relies on the accuracy of the estimated flow, which is still a quite challenging task.
The latter considers pixel synthesis for the interpolated frame as local convolution over input frames and uses a network to predict a convolution kernel for each pixel. However, it has difficulty handling large spatial displacements.

Recently, several methods have applied diffusion models for video interpolation, \textit{e.g.,} LDMVFI~\cite{danier2023ldmvfi}, MCVD~\cite{mcvd}, VIDIM~\cite{jain2024vivdm}, and CBBD~\cite{lyu2024cbbd}.
These methods treat video interpolation as a conditional generation or temporal inpainting task.
Given key frames, the diffusion models are trained to generate intermediate frames through the denoising process.
% MCVD and VIDIM are modeled in pixels space while LDMVFI and CBBD are in latent space. 
In some video generation methods, video interpolation is integrated into the video model as a module, \textit{e.g.,} ImagenVideo, Make-A-Video, LaVie, and Show-1.
The interpolation module is deeply integrated with other modules, which makes it difficult to incorporate the interpolation module into other video models.

Although conventional methods, diffusion-based models, and bounded interpolation modules can generate intermediate frames for constructing a smooth video, all these methods require a training process involving millions or even billions of parameters.
Furthermore, since conventional and diffusion-based models are always trained independently on real data, a domain gap exists when applying them to synthetic videos produced by generative video models.
As for bounded interpolation modules, they need to be retrained if the base video model is updated, and they are not generalizable across different video models.

In this work, we propose a training-free method to improve the frame rate of various existing generative video models, \textit{e.g.,} SVD~\cite{stable-video-diffusion} and VideoCrafter~\cite{chen2023videocrafter1}. 
The proposed method can be applied in a plug-and-play manner across various video models. 
We investigate the temporal correlation of features learned by the video UNet and observe that latent space back-projection fails to inject content and appearance to adjacent frames. Fortunately, we find the temporal correlation learned in transformer hidden states is strong. Back-projection in transformer hidden states achieves strong visual content control while preserving great inter-frame consistency. Based on this observation,
we transform the target video model into a self-cascaded architecture containing two branches (see Fig.~\ref{fig:self-cascaded}).
One branch retains the architecture of the target model for short video inference, while the other is adapted for long video inference by placing the proposed hidden state correction modules into the transformer blocks.
The correction modules use both hidden states from the two branches as input to calibrate the hidden states of the long branch for inter-frame consistency.
Additionally, we design a strategy for the controllability of the correction strength.

Our contributions are summarized as follows:
\begin{itemize}
    \item We propose a training-free method to enhance generative video models to produce videos with a higher frame rate, resulting in visually smooth videos. The method can be applied to various video models in a plug-and-play manner. 
    \item We propose the self-cascaded architecture and the hidden state correction modules for inter-frame consistency. 
    \item We conduct extensive experiments with various popular generative video models to demonstrate the effectiveness of the proposed method, inclusing SVD, VideoCrafter, and LaVie. 
\end{itemize}
\section{Related Works}
\paragraph{Video Diffusion Models} Diffusion models now prevail in the video generation community. They model the video data distribution with a sequence of iterative denoising steps \cite{ddpm, ddim, score-sde}. Diffusion models are used for a variety of video tasks, including action category conditioned video generation \cite{mcvd, VideoGPT, luo2023videofusion, pyoco}, text conditioned video generation \cite{imagen-video, video-ldm, he2022latent}, image conditioned video generation \cite{stable-video-diffusion, yu2023animatezero}, and video translation \cite{make-your-vid, yang2023rerender}. Among them, VideoCrafter\cite{chen2023videocrafter1, chen2024videocrafter2}, LaVie\cite{wang2023lavie}, and StableVideoDiffusion\cite{stable-video-diffusion} approximate the data in a compact video latent space computed by a VAE\cite{VAE}. Meanwhile, Imagen-Video \cite{imagen-video}, Make-A-Video \cite{make-a-video}, and PYoCo \cite{pyoco} learn the pixel space video distribution. Some efforts have been made in long video generation \cite{gen-l-vid, qiu2023freenoise}. Gen-L-Video \cite{gen-l-vid} bootstraps video diffusers to generate beyond the video length limit. FreeNoise \cite{qiu2023freenoise} proposes a noise schedule to alleviate content shifting.

% Diffusion models decompose the generation process into a sequence of progressive denoising steps \cite{ddpm, ddim}. Over the past few years, diffusion models have made significant strides in various tasks including text-to-image generation \cite{imagen, stable-diffusion, sdxl}, text-to-video generation \cite{make-a-video, imagen-video, wang2023lavie}, audio generation \cite{kong2021diffwave}, and 3D generation \cite{poole2022dreamfusion, liu2023zero1to3}. 
% Expanding beyond images, video diffusion models have been introduced to generate videos using a temporal-aware U-Net architecture. 
% This paradigm in both image and video fields can be further categorized into cascaded diffusion models \cite{cascaded-dm} and single-stage diffusion models \cite{stable-diffusion, adm}. 
% Despite their effectiveness, diffusion models have certain limitations, particularly in terms of training efficiency, especially for video generation with a high dimensionality of data samples.

\paragraph{Zero-shot Visual Restoration and Video Interpolation} With the power of modern pre-trained diffusion models, zero-shot visual restoration has made considerable progress \cite{garber2023ddpg, ddnm, chung2023dps, zhu2023diffpir}. ILVR \cite{ilvr} adopts a trained diffuser for zero-shot image super-resolution. DDNM \cite{ddnm} proposes theoretical insights into null-space and range-space decomposition of visual restoration. DDPG \cite{garber2023ddpg} combines the back-projection method and least square method using an optimization preconditioner. However, limited efforts have been made to deliver zero-shot visual restoration to the video domain. ScaleCrafter \cite{he2023scalecrafter} proposes a tuning-free method for inferring at a higher resolution. In this paper, we investigate the potential of zero-shot methods for higher frame rate generation. 

Video interpolation is a long-standing problem in computer vision. Traditional video interpolation models adopt a two-stage paradigm to estimate the optical flow and use it to aid frame interpolation \cite{flow1, flow2, flow4, reda2022film}. Recently, progress has been made in deploying new architectures \cite{lu2022kernel1, zhang2023kernel2} and training paradigms \cite{danier2023ldmvfi, jain2024vivdm, lyu2024cbbd}. VFIformer \cite{lu2022kernel1} adopts a transformer for frame interpolation. LDMVFI\cite{danier2023ldmvfi} and VIVDM\cite{jain2024vivdm} propose training diffusion models for the task.
\section{Preliminary}
\subsection{Video Diffusion Models}
Given a video $\mathbf{x}_0 \sim p(\mathbf{x}_0)$ with $S$ frames, let $ \mathbf{x}_0 = \{\boldsymbol{x}_0^0, \boldsymbol{x}_0^1, \cdots, \boldsymbol{x}_0^S\}$. Generative models approximate a sequence of diffusion denoising transitions \cite{video-dm, ddpm, im-ddpm}
\begin{align}
     q(\mathbf{z}_t \vert \mathbf{x}_0) & = \mathcal{N}(\mathbf{z}_t \vert \alpha_t \mathbf{x}_0, \sigma_t^2 \mathbf{I}) \\
     q(\mathbf{z}_s \vert \mathbf{z}_t, \mathbf{x}_0) & = \mathcal{N}(\mathbf{z}_s \vert \bm{\mu}_{s\vert t}(\mathbf{z}_t, \mathbf{x}_0), \sigma_{s\vert t}^2 \mathbf{I}),
\end{align}
where $\mathbf{z}_t$ is the noisy data at a noise scale $t = 1,2,\cdots,T$ and $0 \leq s < t \leq T$. Let $\lambda_t = \log(\alpha_t^2 / \sigma_t^2)$ be the monotonically decreasing signal-to-noise ratio along noise scale $t$. The mean and variance of backward diffusion process $q(\mathbf{z}_s \vert \mathbf{z}_t, \mathbf{x}_0)$ is
\begin{align}
    \bm{\mu}_{s\vert t}(\mathbf{z}_t, \mathbf{x}_0) & = e^{\lambda_t - \lambda_s} (\frac{\alpha_s}{\alpha_t}) \mathbf{z}_t + (1 - e^{\lambda_t - \lambda_s}) \alpha_s \mathbf{x}_0 \\
    \sigma_{s\vert t}^2 & = (1 - e^{\lambda_t - \lambda_s})\sigma_t^2.
\end{align}

Starting from Gaussian noise $\mathbf{z}_T \sim \mathcal{N}(\mathbf{z}_T \vert \bm{0}, \mathbf{I})$, we train a video diffusion UNet to denoise $\mathbf{z}_T$ iteratively. Specifically, it estimates $p_\theta(\mathbf{z}_s \vert \mathbf{z}_t, y) = \mathcal{N}(\mathbf{z}_s \vert \bm{\mu}_\theta^{s\vert t}(\mathbf{z}_t, y), \bm{\Sigma}_\theta^{s\vert t}(\mathbf{z}_t), y)$, which approximates the reverse diffusion process with condition $y$, i.e. textual prompts. The diffusion UNet $\boldsymbol{\epsilon}_\theta(\cdot)$ is parameterized to estimate the added noise in $\mathbf{x}_t$. One can rearrange the $\boldsymbol{\epsilon}_\theta(\cdot)$ to compute the noiseless prediction $\mathbf{x}_{0\vert t} = \mathbf{f}_\theta(\mathbf{z}_t, y)$ that approximates $\mathbf{x}_0$ \cite{ddim}. 

\subsection{Visual Restoration with Diffusion Models}
The back-projection method \cite{ddnm, garber2023ddpg} is prevalent for zero-shot image restoration. It considers a linear measurement operator $\mathbf{A}$, such that the degenerated sample is $\mathbf{y} = \mathbf{A} \mathbf{x}_0$. Meanwhile, the high-resolution image can be decomposed into two parts
\begin{equation}
    \mathbf{x}_0 = \underbrace{\mathbf{A}^\dagger \mathbf{A} \mathbf{x}_0}_{\mathrm{range space}} + \underbrace{(\mathbf{I} - \mathbf{A}^\dagger \mathbf{A}) \mathbf{x}_0}_{\mathrm{null space}}
\end{equation} 
where $\mathbf{A}^\dagger \mathbf{A} \mathbf{x}_0$ is in the range space of $\mathbf{A}$, $(\mathbf{I} - \mathbf{A}^\dagger \mathbf{A}) \mathbf{x}_0$ is in the null space of $\mathbf{A}$. $\mathbf{A}^\dagger$ is the pseudo inverse of $\mathbf{A}$ such that $\mathbf{A} \mathbf{A}^\dagger \mathbf{A} = \mathbf{A}$. DDNM \cite{ddnm} proposes to back-project the observed part of a sample $\mathbf{y}$ to replace the range space part in denoising process
\begin{equation}
    \hat{\mathbf{x}}_{0 \vert t} = (\mathbf{I} - \mathbf{A}^\dagger \mathbf{A}) \mathbf{x}_{0\vert t} + \mathbf{A}^\dagger \mathbf{y}
\end{equation} where $\hat{\mathbf{x}}_{0\vert t}$ is the corrected estimate. Therefore, the diffusion model only predicts in the null space of $\mathbf{A}$ and ensures the consistency to condition $\mathbf{y}$ in the output.
\section{Method}
\begin{figure*}[tb]
  \centering
  \includegraphics[width=1.0\textwidth]{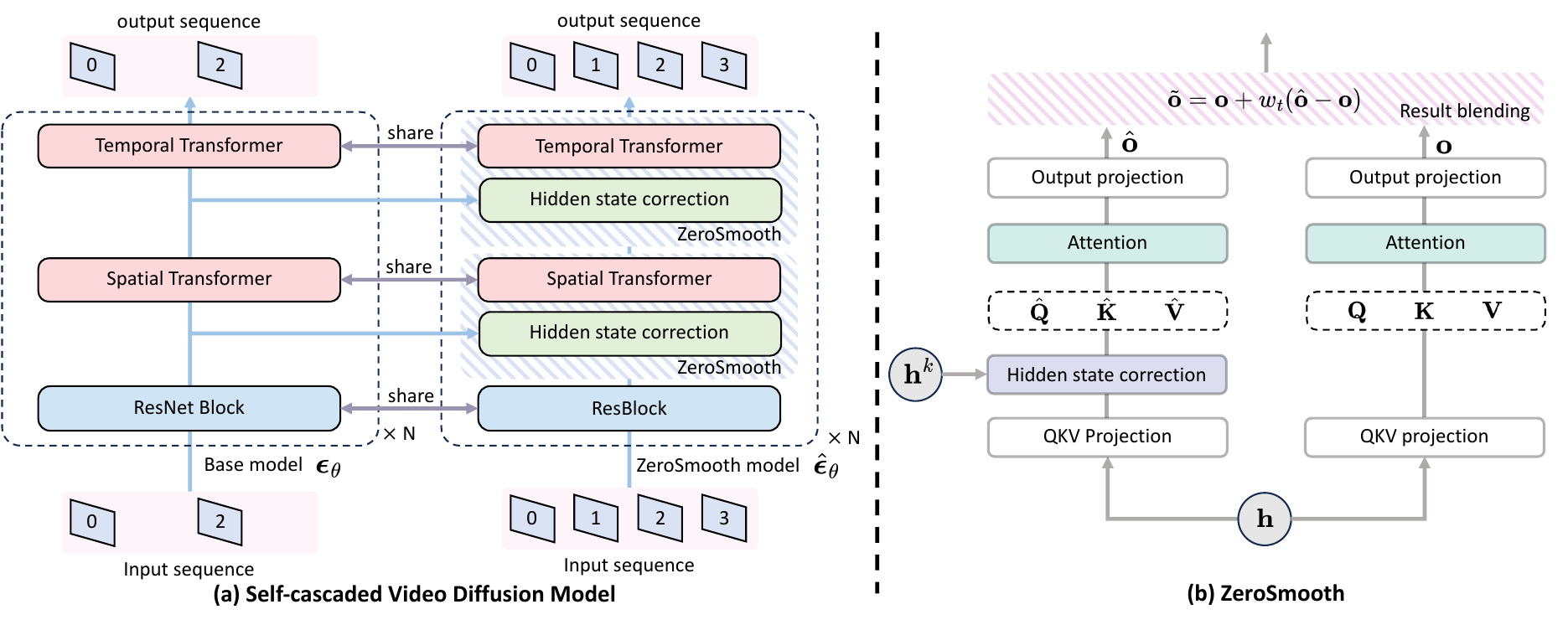}
  \caption{
    An overview of our method. \textbf{(a)} We build cascaded video diffusion model by adapting the base generator to generate at a higher frame rate. \textbf{(b)} A sketch for hidden states correction in transformers in ZeroSmooth.
  }
  \label{fig:self-cascaded}
\end{figure*}
We aim to adapt a video diffusion model to generate at a higher frame rate while preserving its original generation capability. Given a pretrained video diffusion model $\epsilon_\theta(\cdot): \mathbb{R}^{c \times t_0 \times h \times w} \rightarrow \mathbb{R}^{c \times t_0 \times h \times w}$,  which is trained on $t_0$ frame videos. Our method creates $\hat{\epsilon}_\theta(\cdot): \mathbb{R}^{c \times t \times h \times w} \rightarrow \mathbb{R}^{c \times t \times h \times w}$ from $\epsilon_\theta(\cdot)$ in a training-free manner, where $t > t_0$. 
% We ensure the generated video is viasually plausible and follows the content and appearance of a base video generated by $\hat{\boldsymbol{\epsilon}}_\theta(\cdot)$.
Once the key frames $\mathbf{y}$ are observed, generating the corresponding high frame rate video can be considered video frame interpolation (VFI). Meanwhile, the measurement operator $\mathbf{A}^{t_0 \times t}$, which acquires key frames from a full video $\mathbf{x}_0$ is linear and follows the formulation of $\mathbf{y} = \mathbf{A} \mathbf{x}_0$.  An intuitive idea to achieve this is to apply DDNM \cite{ddnm} to the $t_0$-frame base video generated by $\epsilon_\theta(\cdot)$, interpolating between the key frames and get the $t$-frame high frame rate video. During this process, one can build a $t$-frame generator $\hat{\epsilon}_\theta(\cdot)$ using the method proposed by Qiu et al. \cite{qiu2023freenoise} After each denoising step, the key frames are back-projected to correct the estimate. 

However, as seen in \cref{fig:compare-tuning-free}, we find it challenging for this intuitive approach to maintain temporal consistency between observed and interpolated frames. We attribute this to the weak temporal correlation of the video latent space learned by diffusers. Correcting the estimate in the latent space can hardly inject content and appearance into the interpolated frames.
% If $\hat{\epsilon}_\theta(\cdot)$ approximates the data distribution well, then $\mathbf{A}^\dagger \mathbf{y}$ and  $(\mathbf{I} - \mathbf{A}^\dagger \mathbf{A})\mathbf{x}_{0\vert t}$ satisfy the marginal data distribution in the range-space and null-space respectively. Unfortunately, when combining them together. $\hat{\mathbf{x}}_{0|t}$ fails to match the joint distribution 
Alternatively, we propose to correct the hidden states of transformer modules where the temporal correlation is strong. Our method is summarized in three parts. 
In the following subsections, we introduce the overall architecture of our method, the hidden state correction method in transformers, and our remedy to alleviate joint distribution mismatch after correction.

% Without loss of generality, we show the situation of 2 times higher frame rate 

\subsection{Self-cascaded Video Diffusion Model}
As shown in \cref{fig:self-cascaded}(a), we construct a cascaded video diffusion model consisting of $\epsilon_\theta(\cdot)$ and $\hat{\epsilon}_\theta(\cdot)$. In our method, $\epsilon_\theta(\cdot)$ and $\hat{\epsilon}_\theta(\cdot)$ share the same neural network modules and parameters. Therefore, we call this paradigm a \emph{self-cascaded video diffusion model}. In order to correct the high frame rate estimate, we first generate key frame transformer hidden states.
The hidden state distribution in a diffusion model varies along denoising timesteps. Instead of using only a final key frame video for back-projection like DDNM, our method uses the key frame hidden states at the corresponding step to match the data distribution at different timesteps. 

Mathematically, we denote the input to spatial or temporal transformer as $\mathbf{h} \in \mathbb{R}^{t \times l \times d}$ in $\hat{\boldsymbol{\epsilon}}_\theta(\cdot )$. $l$ is the stacked spatial dimension (i.e., feature width times feature height), and $d$ is the hidden state channel. The hidden state input for the corresponding transformer in $\boldsymbol{\epsilon}_\theta(\cdot)$ is $\mathbf{h}^k \in \mathbb{R}^{t_0 \times l \times d}$. As shown in \cref{fig:self-cascaded}(b), we apply back-projection to correct the hidden state estimation with key frame hidden states $\mathbf{h}^k$ before computing attention. We denote the query, key and value with $\mathbf{Q} \in \mathbb{R}^{t \times l \times d}, \mathbf{K}\in \mathbb{R}^{t\times l \times d}, \mathbf{V} \in \mathbb{R}^{t \times l \times d}$. After hidden state correction, we use $\hat{\mathbf{Q}}, \hat{\mathbf{K}}$ and $\hat{\mathbf{V}}$ to represent them. Note that the self-cascaded video diffusion model can include multiple stages to achieve higher frame rate results. For instance, consider $t = nt_0$, where $n$ is the video interpolation scale. One can construct another model $\tilde{\epsilon}_\theta(\cdot): \mathbb{R}^{c \times n^2 t_0 \times h \times w} \rightarrow \mathbb{R}^{c \times n^2 t_0 \times h \times w}$ from $\epsilon_\theta(\cdot)$. The self-cascaded model $\{\epsilon_\theta, \hat{\epsilon}_\theta, \tilde{\epsilon}_\theta\}$ achieves $n^2$ times more frames than the key frame generator.

\paragraph{Temporal Attention for High Frame Rate Generation}
A pretrained video diffusion model is trained on videos of length $t_0$. Therefore, $t$-frame video is unseen for the temporal transformer modules. Direct inference without adaptation will result in severe sample quality degradation. We propose \emph{ZeroSmooth temporal attention} to adapt the modules for a longer sequence. Video diffusers use different categories of temporal attention. Text-to-video diffusion models, like LaVie\cite{wang2023lavie} and VideoCrafter1 \cite{chen2023videocrafter1}, use relative positional embedding (RPE). The sequence length should be kept unchanged when processing high frame rate input. We improve upon the attention fusion technique in FreeNoise \cite{qiu2023freenoise} by adding RPE to every attention window, ensuring correct relative time perception. Image-to-video diffusion models, like StableVideoDiffusion \cite{stable-video-diffusion}, use absolute positional embedding (APE) to provide temporal distance between the current frame and the reference image. Therefore, we interpolate the position index to maintain the temporal distance in the high frame rate video.
% Text-to-video diffusers and image-to-video diffusers 

\subsection{Hidden State Correction for Transformers}
We adopt back-projection method in transformer modules of video diffusion to correct hidden states estimate. Without loss of generality, we consider a $2\times$ frame interpolation in this subsection. The measurement for video frame interpolation can be defined in two ways. First, we can use sampling operator $\mathbf{A} \in \mathbb{R}^{t_0 \times 2t_0}$ to draw key frames from a full video. Second, one can use the interpolation operator $\{\mathbf{A}_1, \mathbf{A}_2\}$ to perform temporal downsampling to get key frames, in which the align corners are different in $\mathbf{A}_1$ and $\mathbf{A}_2$. The measurement matrices are shown below for the $2\times$ interpolation case:
\begin{align}
    \mathbf{A} = \left[\begin{matrix}
        1 & 0 & 0 & 0 & \cdots & 0 & 0 \\
        0 & 0 & 1 & 0 & \cdots & 0 & 0 \\
        \vdots & \vdots & \vdots & \vdots & & \vdots & \vdots \\
        0 & 0 & 0 & 0 & \cdots & 1 & 0
    \end{matrix}\right] \qquad \qquad \qquad \qquad \qquad \quad \\
    \mathbf{A}_1 = \left[\begin{matrix}
        0.5 & 0.5 & 0 & 0 & \cdots & 0 & 0 \\
        0 & 0 & 0.5 & 0.5 & \cdots & 0 & 0 \\
        \vdots & \vdots & \vdots & \vdots & & \vdots & \vdots \\
        0 & 0 & 0 & 0 & \cdots & 0.5 & 0.5
    \end{matrix}\right], \mathbf{A}_2 = \left[\begin{matrix}
        1 & 0 & 0 & 0 & \cdots & 0 & 0 \\
        0 & 0.5 & 0.5 & 0 & \cdots & 0 & 0 \\
        \vdots & \vdots & \vdots & \vdots & & \vdots & \vdots \\
        0 & 0 & 0 & 0 & \cdots & 0.5 & 0 
    \end{matrix}\right].
\end{align}

\begin{figure*}[tb]
  \centering
  \includegraphics[width=1.0\textwidth]{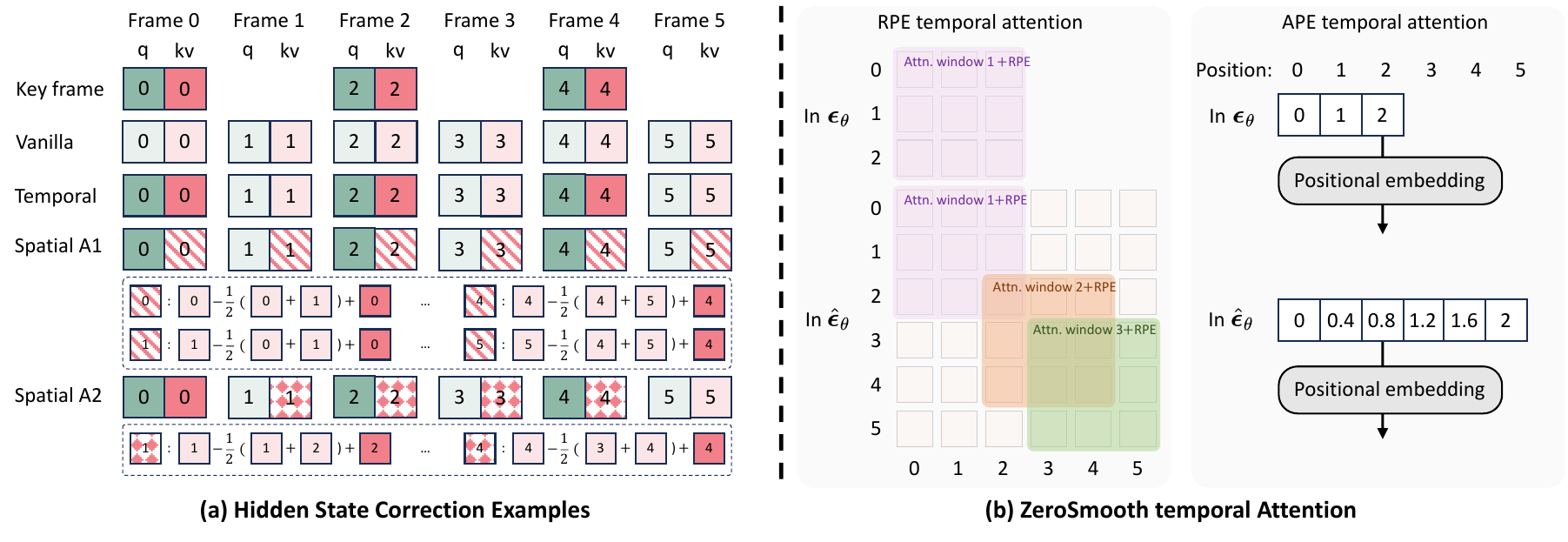}
  \caption{
    \textbf{(a)} Examples for hidden states correction in 2$\times$ higher frame rate generation case, showcasing the queries, keys and values are calibration in temporal transformer (Temporal), and in spatial transformers using different interpolation operators (Spatial A1, Spatial A2). \textbf{(b)} We adapt temporal transformers to generate longer sequences in different ways. For the temporal module with relative positional embedding (RPE), we use windowed attention and apply RPE within each window. For absolute positional embedding (APE) modules, we interpolate the position index to get APE before applying attention operation.
  }
  \label{fig:backprojection}
\end{figure*}

Intuitively, sampling operator $\mathbf{A}$ is less likely to cause blur in key frames since it directly copies corresponding frames. However, we find $\mathbf{A}_1$ and $\mathbf{A}_2$ especially useful for the key and value in spatial transformer modules. Within these modules, the query determines the scene structure of a frame while the key and value provide textures and appearances. As a result, using interpolation operator for back-projection in the spatial transformer key and value offers additional temporal consistency in visual details without causing blurry scene structures. Examples of hidden state correction is shown in \cref{fig:backprojection}(a). Mathmatically, the correction for temporal transformer hidden states is
\begin{equation}
    \hat{\mathbf{h}}^{\mathrm{temp}} = (\mathbf{I} - \mathbf{A}^\dagger \mathbf{A})\mathbf{h}^{\mathrm{temp}} + \mathbf{A}^\dagger \mathbf{h}^k,
\end{equation} where $\mathbf{h}^{\mathrm{temp}}$ and $\hat{\mathbf{h}}^{\mathrm{temp}}$ are the temporal hidden states and the corrected ones. In spatial transformers, the correction is
\begin{align}
    & \qquad \qquad \qquad \qquad \qquad \hat{\mathbf{Q}}^{\mathrm{spatial}} = (\mathbf{I} - \mathbf{A}^\dagger \mathbf{A})\mathbf{Q}^{\mathrm{spatial}} + \mathbf{A}^\dagger \mathbf{W}_q \mathbf{h}^k & \\
    & \qquad \qquad \qquad \qquad \qquad \hat{\mathbf{K}}^{\mathrm{spatial}} = \mathbbm{1}_{p > 0.5} \hat{\mathbf{K}}^{\mathrm{spatial}}_1 + (1 - \mathbbm{1}_{p > 0.5}) \hat{\mathbf{K}}^{\mathrm{spatial}}_2 \\
    & \qquad \qquad \qquad \qquad \qquad \hat{\mathbf{V}}^{\mathrm{spatial}} = \mathbbm{1}_{p > 0.5} \hat{\mathbf{V}}^{\mathrm{spatial}}_1 + (1 - \mathbbm{1}_{p > 0.5}) \hat{\mathbf{V}}^{\mathrm{spatial}}_2 \\
    & \qquad \hat{\mathbf{K}}^{\mathrm{spatial}}_1  = (\mathbf{I} - \mathbf{A}_1^\dagger \mathbf{A}_1)\mathbf{K}^{\mathrm{spatial}} + \mathbf{A}_1^\dagger \mathbf{W}_k \mathbf{h}^k, \hat{\mathbf{K}}^{\mathrm{spatial}}_2 = (\mathbf{I} - \mathbf{A}_2^\dagger \mathbf{A}_2)\mathbf{K}^{\mathrm{spatial}} + \mathbf{A}_2^\dagger \mathbf{W}_k \mathbf{h}^k \\
    & \qquad \hat{\mathbf{V}}^{\mathrm{spatial}}_1  = (\mathbf{I} - \mathbf{A}_1^\dagger \mathbf{A}_1)\mathbf{V}^{\mathrm{spatial}} + \mathbf{A}_1^\dagger \mathbf{W}_v \mathbf{h}^k, \hat{\mathbf{V}}^{\mathrm{spatial}}_2 = (\mathbf{I} - \mathbf{A}_2^\dagger \mathbf{A}_2)\mathbf{V}^{\mathrm{spatial}} + \mathbf{A}_2^\dagger \mathbf{W}_v \mathbf{h}^k.
\end{align} 
where $p$ is a random value drawn from standard Gaussian. $\mathbbm{1}_{p > 0.5}$ equals 1 if $p > 0.5$ and equals 0 otherwise. We randomly use $\mathbf{A}_1$ and $\mathbf{A}_2$ to correct $\mathbf{K}$ and $\mathbf{V}$, since using only a single interpolation function will cause biased time stamp in interpolated frames (i.e. the 2-nd frame will look like 1.5-th frame if use $\mathbf{A}_1$ only) due to the align corner.
% the hidden state correction in transformers can be summarized as follows:

\subsection{Controlling Correction Strength}
% \paragraph{Joint distribution problem of back-projection}
In vanilla DDNM, if $\hat{\epsilon}_\theta(\cdot)$ approximates the data distribution well, then $\mathbf{A}^\dagger \mathbf{y}$ and  $(\mathbf{I} - \mathbf{A}^\dagger \mathbf{A})\mathbf{x}_{0\vert t}$ satisfy the marginal data distribution in the range-space and null-space respectively. Unfortunately, when combining them together. $\hat{\mathbf{x}}_{0|t}$ may fail to match the joint distribution. This problem also presents in our transformer hidden state back-projection. We propose a simple yet effective remedy for this shown in \cref{fig:self-cascaded}(b). Instead of using the output $\hat{\mathbf{o}}$ from corrected hidden states alone. Our method utilize a linear interpolation of the original output $\mathbf{o}$ and the corrected output $\hat{\mathbf{o}}$ as the final result. Mathmatically, the controlled corrected output is
\begin{equation}
    \tilde{\mathbf{o}} = \mathbf{o} + w_t (\hat{\mathbf{o}} - \mathbf{o}).
    \label{eqn:control}
\end{equation} $w_t$ is a predefined control scale that lies between 0 and 1 for denoising timestep $t$. With Eqn.\ref{eqn:control}, we are able to control how much a denoising step relies on the corrected output. By introducing $\mathbf{O}$ to the final result, we alleviate the artifact caused by the joint distribution mismatch of $\hat{\mathbf{h}}$.

\section{Experiments}

\paragraph{Experiment Settings} We deploy our method on three prevalent video diffusers including VideoCrafter2\cite{chen2024videocrafter2}, LaVie\cite{wang2023lavie} and StableVideoDiffusion\cite{stable-video-diffusion}. The base model of VideoCrafter2 and LaVie generates 16 frames 320$\times$512 videos using textual prompt. StableVideoDiffusion generates 14 frames 576$\times$1024 videos given a reference start frame. The testing tasks includes text-to-video generation and image-to-video generation. Specifically, we use self-cascaded VideoCrafter2 and LaVie for text-to-video generation, and use self-cascaded StableVideoDiffusion for image-to-video. For all the experiments, we use our method to inference in two untrained frame rate settings including 2$\times$ and $4\times$ higher frame rate without any further model tuning. The noise in key frame is set to be identical to the base generation stage in every timesteps to alleviate the stochasticity of denoising. Please see the detailed inference hyperparameters in our appendix.

\paragraph{Testing Datasets and Metrics}

% hesis evaluation protocol of previous works. The evaluations are conducted in two datasets UCF-101 \cite{UCF101} and MSR-VTT \cite{msrvtt}. UCF-101 is an action recognition dataset that contains categorized videos in 101 motion classes. We measure Frechet Video Distance (FVD) and Inception Score (IS) on the dataset. To generate videos with the UCF-101 action class using a text-to-video model, we use the text prompt constructed by Ge et al.\cite{pyoco}. For the text prompt for each class, we sample 20 videos, getting 2,020 videos in total.

For text-to-video generation, we use UCF-101\cite{UCF101} as the testing dataset. UCF-101 is a video dataset that includes 101 action categories. To use it as a text-to-video benchmark, we follow the text prompt proposed by Gu et al.\cite{pyoco} For each category of action, we generate 20 videos and getting 2,020 videos in total for every method. For both VideoCrafter2 and LaVie, we generate 320$\times$ 512 videos with 16, 32 and 64 frames. VideoCrafter2 uses a frame-per-second condition and we set it to 8, 16, and 32 respectively. The sample quality is evaluated with Inception Score (IS) and Frechet Video Distance (FVD). The consistency to key frame condition is measured by SSIM and PSNR. To get these metrics, we compute the image similarity between key frames in the high frame rate video and the base video. For image-to-video generation, we use WebVid-10M\cite{webvid} as the testing dataset. It is a high definition video dataset within the domain of StableVideoDiffusion. WebVid-10M contains over 10 million videos in various scenarios. We randomly sample 2,048 video clips from the dataset and use the first frame of each video to serve as reference image. We use StableVideoDiffusion to generate $576\times 1024$ videos with 14, 28 and 56 frames. The motion strength is 180, and the frame-per-second is set to 7 in the model. For every method in image-to-video, we generate 2,048 videos. FVD, SSIM and PSNR are used to evaluate the video quality and the similarity of key frames compared to the base video condition. 

\subsection{Comparison to Tuning-free Methods}
\begin{table}[t]
    \centering
    \begin{tabular}{cccccccccccccc}
    \toprule
        \multirow{2}{*}{Model} & \multirow{2}{*}{Method} & \multicolumn{4}{c}{2$\times$} & \multicolumn{4}{c}{4$\times$} \\
        \cmidrule(lr){3-6} \cmidrule(lr){7-10}
        % \midrule
        & & FVD$\downarrow$ & IS$\uparrow$ & PSNR$\uparrow$ & SSIM$\uparrow$ & FVD$\downarrow$ & IS$\uparrow$ & PSNR$\uparrow$ & SSIM$\uparrow$ \\
        \midrule
        % 1:1, 1$\times$ & & xxx & xxx & xxx & xxx & 12.07 & 0.004 & - & - & xxx & xxx & xxx & xxx \\
        % \midrule
        %
        \multirow{3}{*}{VC2} & Dir. Inf. & 1202.3 & 30.31 & 10.90 & 0.426 & 1183.9 & 21.63 & 10.81 & 0.409  \\
        & DDNM & 1139.6 & 40.99 & 23.50 & 0.841 & 1100.4 & 35.90 & 22.21 & 0.813 \\
        & Ours & \textbf{1085.8} & \textbf{43.92} & \textbf{24.12} & \textbf{0.852} & \textbf{1020.6} & \textbf{43.77} & \textbf{22.68} & \textbf{0.837} \\
        \midrule
        \multirow{3}{*}{LaVie} & Dir. Inf. & \textbf{682.0} & 28.11 & 10.21 & 0.438 & 834.7 & 24.23 & 10.19 & 0.447  \\
        & DDNM & 729.5 & 35.59 & 24.39 & 0.824 & 721.0 & 29.42 & \textbf{26.47} & 0.817  \\
        & Ours & 703.2 & \textbf{39.23} & \textbf{26.90} & \textbf{0.876} & \textbf{697.6} & \textbf{38.67} & 24.17 & \textbf{0.831} \\
        \midrule
        \multirow{3}{*}{SVD} & Dir. Inf. & 1350.7 & - & 17.33 & 0.651 & 1090.2 & - & 15.75 & 0.604 \\
        & DDNM & 846.6 & - & 29.12 & 0.875 & 895.2 & - & 28.12 & 0.864 \\
        & Ours & \textbf{779.6} & - & \textbf{30.92} & \textbf{0.927} & \textbf{752.1} & - & \textbf{30.74} & \textbf{0.921} \\
        \bottomrule
    \end{tabular}
    \vspace{5pt}
    \caption{
    Quantitative comparisons between tuning-free methods to generate 2$\times$ and 4$\times$ higher frame rate.
    }
    \label{tab:compare-tuning-free}
    % \vspace{-10pt}
\end{table}
% \begin{figure*}[tb]
%   \centering
%   \includegraphics[width=1.00\textwidth]{res/compare-tuning-free.pdf}
%   \caption{
%     Qualitative comparison bettween our method, Gen-2 \cite{gen-1}, VideoCrafter1 \cite{chen2023videocrafter1} and ModelScope \cite{modelscope}.
%    }
%   \label{fig:compare-funing-free}
% \end{figure*}

\begin{figure}[tb]
    \makebox[\textwidth][c]{
    \includegraphics[width=1.2\textwidth]{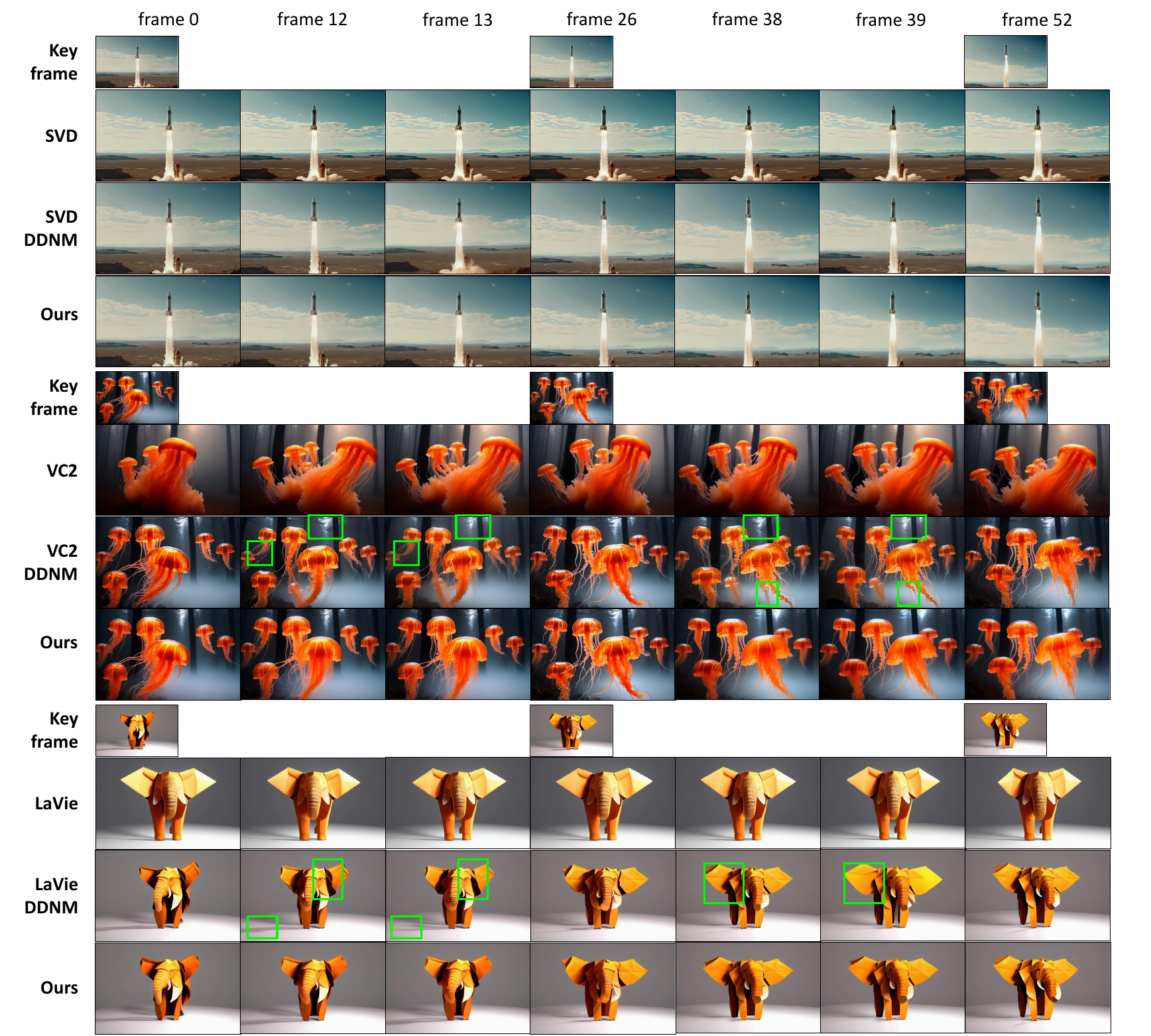}}  
    \caption{Visual comparison between our method and other tuning-free baselines. The green rectangles capture the abrupt changes between adjacent frames.} 
    \label{fig:compare-tuning-free}
\end{figure}

We compare our method to the vanilla model (Direct inference), a tuning-free method for visual restoration via back-projection in latent space named DDNM \cite{ddnm}. Our results are shown in Tab.\ref{tab:compare-tuning-free}. A qualitative comparison is shown in Fig.\ref{fig:compare-tuning-free}. For the best visual experience, please see the videos in our supplementary material. We surpass the baselines in most of the metrics. The similarity metrics PSNR and SSIM show our method is able to preserve key frame when generating in high frame rate. Meanwhile, our method achieves the best FVD and IS. ZeroSmooth generates visual plausible and temporal consistency high frame rate videos. We show the method is effective in VideoCrafter2, LaVie and StableVideoDiffusion. It illustrates our method's potential to be an model agnostic method and can be generalized to other video generative diffusers. 

 The result of DDNM shows it is hard to apply back-projection in video latent space for temporal consistent video interpolation. Generating higher frame rate video directly is out of the domain of a pretrained video diffuser. Therefore inference directly deteriorates in visual quality and text-video alignment. Due to the stochasticity introduced by interpolated frames. Direct inference fails to maintain key frame contents even when the key frame noise is identical to that during the base video generation.
% \subsection{Comparison to video diffusers}
\subsection{Comparison to Training-based Methods}
\begin{table}[t]
    \centering
    \begin{tabular}{ccccccccc}
    \toprule
        \multirow{2}{*}{Method} & \multicolumn{4}{c}{2$\times$} & \multicolumn{4}{c}{4$\times$} \\
        \cmidrule(lr){2-5} \cmidrule(lr){6-9}
        % \midrule
        & FVD$\downarrow$ & IS$\uparrow$ & PSNR$\uparrow$ & SSIM$\uparrow$ & FVD$\downarrow$ & IS$\uparrow$ & PSNR$\uparrow$ & SSIM$\uparrow$ \\
        \midrule
        % 1:1, 1$\times$ & & xxx & xxx & xxx & xxx & 12.07 & 0.004 & - & - & xxx & xxx & xxx & xxx \\
        % \midrule
        %
        % SGM-VFI & 38.50 & 0.014 & 29.30 & 0.008 & 29.89 & 0.010 & 24.21 & 0.007  \\
        LDMVFI & 1304.6 & 41.40 & \textbf{34.60} & \textbf{0.948} & - & - & - & - \\
        LaVie Intp. & - & - & - & - & 1172.1 & 35.78 & \textbf{28.44} & \textbf{0.874} \\
        Ours & \textbf{1085.8} & \textbf{43.92} & 24.12 & 0.852 & \textbf{1020.6} & \textbf{43.77} & 22.68 & 0.837 \\
        \bottomrule
    \end{tabular}
    \vspace{5pt}
    \caption{
    Quantitative comparisons between our method and training-based video interpolators.
    }
    \label{tab:compare-interpolation}
    % \vspace{-10pt}
\end{table}
Despite ZeroSmooth is zero-shot method and no additional tuning or training data is required. We find our method competitive with those training-based video interpolation models. We compare our method with video interpolation models including LDMVFI\cite{danier2023ldmvfi} and LaVie interpolation model \cite{wang2023lavie}. Since our method is aimed for text-to-video and image-to-video generation, but not for real world video interpolation. We use the methods to interpolate the generated videos of VideoCrafter2 base generator for a fair comparison. The results are shown in Tab.\ref{tab:compare-interpolation}. Our method surpasses training-based method in visual quality and is competitive in content preserving. Please see more visual comparisons in our supplementary material.

\subsection{Ablation Study}
\begin{table}[t]
    \centering
    \begin{tabular}{lcccccccc}
    \toprule
        \multirow{2}{*}{Setting} & \multicolumn{4}{c}{2$\times$} & \multicolumn{4}{c}{4$\times$} \\
        \cmidrule(lr){2-5} \cmidrule(lr){6-9}
        % \midrule
        & FVD$\downarrow$ & IS$\uparrow$ & PSNR$\uparrow$ & SSIM$\uparrow$ & FVD$\downarrow$ & IS$\uparrow$ & PSNR$\uparrow$ & SSIM$\uparrow$ \\
        \midrule
        % 1:1, 1$\times$ & & xxx & xxx & xxx & xxx & 12.07 & 0.004 & - & - & xxx & xxx & xxx & xxx \\
        % \midrule
        %
        Vanilla & 1202.3 & 30.31 & 10.90 & 0.426 & 1183.9 & 21.63 & 10.81 & 0.409   \\
        + Spatial & 1251.2 & 42.40 & 22.96 & 0.739 & 1116.8 & 41.19 & 22.31 & 0.721 \\
        + Temopral & 1221.0 & 43.25 & 24.03 & 0.826 & 1156.1 & 42.54 & 22.63 & 0.817 \\
        + Spatial A1/A2 & 1199.8 & 43.09 & 23.22 & 0.768 & 1108.4 & 42.43 & 22.14 & 0.755 \\
        + CCS & \textbf{1085.8} & \textbf{43.95} & \textbf{24.12} & \textbf{0.852} & \textbf{1020.6} & \textbf{43.77} & \textbf{22.68} & \textbf{0.837} \\
        \bottomrule
    \end{tabular}
    \vspace{5pt}
    \caption{
    Quantitative results of ablation experiments.
    }
    \label{tab:ablation}
    % \vspace{-10pt}
\end{table}

% Vanilla & 38.50 & 30.31 & 29.30 & 0.008 & 29.89 & 21.63 & 24.21 & 0.007  \\
%         + Spatial & ? & 43.09 & ? & 0.948 & ? & 42.43 & ? & ? \\
%         + Temopral & ? & 43.92 & ? & ? & ? & 42.54 & ? & ? \\
%         + Spatial A1/A2 & ? & 43.25 & ? & ? & ? & 43.77 & ? & ? \\
%         + CCS & \textbf{1085.8} & \textbf{42.40} & \textbf{24.12} & \textbf{0.852} & \textbf{1086.2} & \textbf{41.19} & \textbf{22.68} & \textbf{0.837} \\
% \input{fig/fig_ablation}
We conduct an ablation experiment of our method on VideoCrafter2 to generate $320\times 512$ videos with 32 and 64 frames. It includes three crucial technical components of our method, including spatial transformers hidden state correction (+ Spatial), spatial hidden state correction with interpolation operator (+ Spatial A1/A2), temporal hidden state correction (+ Temporal), and controlling correction strength (+ CCS). The result is shown in Tab.\ref{tab:ablation}. 
% A visual comparison is in our supplementary material. 
It shows the technical components of ZeroSmooth ensures high quality and content preserving high frame rate video generation. Without the key designs of our method, the visual quality and consistency to key frame content degenerate dramatically.
% can bring about significant improvement in video quality and consistency to the key frame appearance. 
% It shows spatial transformer preserves the visual content of key frames. Spatial hidden states with interpolation operator and temporal hidden state correction enhances temporal consistency. Controlling correction strength improves video smoothness and visual quality. 
% The technical designs in ZeroSmooth ensures high quality and high frame rate video generation.

\section{Conclusion}
We propose a training-free method for promoting generative video diffusion models to generate videos with a high frame rate, \textit{i.e.,} producing more smooth videos.  It can be applied to different video models in a plug-and-play manner.  
Since most video models are trained on videos consisting of sampled key frames due to limited GPU memory, their performance will degenerate sharply if generating a video with more frames than those used in training.  
We design the self-cascaded framework and the hidden state correction module to avoid the degeneration and achieve temporal consistency between frames. 
Extensive experiments are conducted with various video models to demonstrate the effectiveness of the proposed method. 

% We explore the a zero-shot adaptation method to use a pretained video diffuser for high frame rate video generation while preserving the content of the base video. Due to the stochastic nature of denoising iteration and training domain limitation, directly inference a diffuser can hardly keep the visual content and ensure visual quality. We extend the back-projection paradigm in visual restoration to video interpolation and propose back-projection correction of transformer hidden states, which successfully bootstrap the massive prior knowledge of a video diffuser to generate high frame rate videos. Experiments show our method is promising and is potential to generalize to other video diffusers.

\paragraph{Limitations} The proposed method is designed for generative video diffusion models. 
The interpolation performance heavily depends on the capability of the target video model in frame consistency and visual quality.  
If the generated key frames are inconsistent or blurry, our method will fail to improve the video smoothness.  

\paragraph{Boarder Impact}
The proposed interpolation method have a broad impact, enhancing video quality and enabling applications like slow-motion effects, and video compression. They're beneficial in industries like virtual reality, gaming, and film production. However, they also raise ethical concerns, such as the potential for creating misleading deepfake videos, necessitating their responsible use.

% \section{Limitations and Boarder Impact}
% \section*{References}

{
    \small
    \bibliographystyle{splncs04}
    \bibliography{main}
}

% References follow the acknowledgments in the camera-ready paper. Use unnumbered first-level heading for
% the references. Any choice of citation style is acceptable as long as you are
% consistent. It is permissible to reduce the font size to \verb+small+ (9 point)
% when listing the references.
% Note that the Reference section does not count towards the page limit.
% \medskip

% {
% \small

% [1] Alexander, J.A.\ \& Mozer, M.C.\ (1995) Template-based algorithms for
% connectionist rule extraction. In G.\ Tesauro, D.S.\ Touretzky and T.K.\ Leen
% (eds.), {\it Advances in Neural Information Processing Systems 7},
% pp.\ 609--616. Cambridge, MA: MIT Press.

% [2] Bower, J.M.\ \& Beeman, D.\ (1995) {\it The Book of GENESIS: Exploring
%   Realistic Neural Models with the GEneral NEural SImulation System.}  New York:
% TELOS/Springer--Verlag.

% [3] Hasselmo, M.E., Schnell, E.\ \& Barkai, E.\ (1995) Dynamics of learning and
% recall at excitatory recurrent synapses and cholinergic modulation in rat
% hippocampal region CA3. {\it Journal of Neuroscience} {\bf 15}(7):5249-5262.
% }

%%%%%%%%%%%%%%%%%%%%%%%%%%%%%%%%%%%%%%%%%%%%%%%%%%%%%%%%%%%%

\newpage
\appendix

\section{Appendix}
\subsection{ZeroSmooth Temporal Attention}
Detailed computation of ZeroSmooth temporal attention is shown in this subsection. As shown in the main text, we have different strategies for temporal attention modules with relative positional embedding (RPE) and absolute positional embedding (APE). Specifically, in our experiment, LaVie uses RPE in temporal transformers. StableVideoDiffusion uses APE in temporal transformers. VideoCrafter2 learns relative position information within its temporal convolution blocks and no positional embedding is used in temporal transformers. We treat VideoCrafter2 as it has empty RPE. 

\paragraph{Absolute positional embedding} We interpolate the position index before computing absolute positional embedding. In the base video model, the video length is $\mathbf{t_0}$. Consider a adapted model that generates $t$ frames. StableVideoDiffusion uses sinusoidal absolute postition embedding. Mathmatically, the APE is then interpolated 
\begin{align}
     \mathrm{APE}_{(\mathrm{pos}, 2i)}      & = \mathrm{sin}\left(\frac{\mathrm{pos} \cdot t_0}{10000^{2i/d} \cdot t} \right), \\
     \mathrm{APE}_{(\mathrm{pos}, 2i+1)}    & = \mathrm{cos}\left(\frac{\mathrm{pos} \cdot t_0}{10000^{2i/d} \cdot t} \right).
\end{align} $\mathrm{pos} \in \{1, 2, \cdots, t\}$ is the position of the current token. Then a linear mapping layer is applied to get the final absolute positional embedding $\mathbf{U} \in \mathbb{R}^{t \times d}$. The APE is added to hidden state $\mathbf{h}$ before attention operation. We then use windowed attention where the window size equals to sequence length during model training to maintain the perception field of attention operation. Our method ensures there are overlaps between attention windows, such that we can apply the attention fusion technique proposed by Qiu et al\cite{qiu2023freenoise}. to achieve smooth transition between adjacent attention windows. We use an overlap of $\lfloor t_0 / 2 \rfloor$ in our experiment. The the $i$-th attention window starts at position $i_\mathrm{start} = \mathrm{min}(i t_0, t - t_0)$ and ends at $i_\mathrm{end} = \mathrm{min}((i+1)t_0, t)$. We use the subscript $i$ to denote the tokens within position $i_\mathrm{start}$ and $i_\mathrm{end}$. The output of the $i$-th attention window is:
\begin{align}
    \mathbf{o}_i & = \mathrm{Attention}(\mathbf{h}_i + \mathbf{U}_i) \\
                 & = \mathrm{Softmax}\left(\frac{\mathbf{W}_q (\mathbf{h}_i + \mathbf{U}_i)(\mathbf{h}_i + \mathbf{U}_i)^T \mathbf{W}_k^T}{\sqrt{d}}\right) \mathbf{W}_v (\mathbf{h}_i + \mathbf{U}_i),
\end{align} where $\mathbf{W}_q, \mathbf{W}_k, \mathbf{W}_v$ are the projection weight for query, key and value. The final output is
\begin{equation}
    \mathbf{o} = \mathrm{AttentionFusion}(\mathbf{o}_1, \mathbf{o}_2, \cdots, \mathbf{o}_n).
\end{equation} $\mathrm{AttentionFusion}(\cdot)$ is the attention fusion operator in \cite{qiu2023freenoise}. $n$ is the number of windows, one can compute it through $n = \lceil (t - t_0) / (t_0 - \lfloor t_0 /2 \rfloor)\rceil$, where $\lceil \cdot \rceil$ is the ceil operator.

\paragraph{Relative positional embedding} There are many forms of RPEs. Without loss of generality, we consider trainable relative positional embedding in the followings. Let $\mathbf{p}_k \in \mathbb{R}^{t_0 \times d}$ and $\mathbf{p}_v \in \mathbb{R}^{t_0 \times d}$ denote the learned relative position. Then we also apply the windowed attention and attention fusion strategy. Mathmatically, the attention output is
\begin{align}
    \mathbf{o}_i & = \mathrm{Softmax}\left(\frac{\mathbf{W}_q \mathbf{h}_i(\mathbf{h}_i + \mathbf{p}_k)^T\mathbf{W}_k}{\sqrt{d}}\right)\mathbf{W}_v (\mathbf{h}_i + \mathbf{p}_v) \\
    \mathbf{o}   & = \mathrm{AttentionFusion}(\mathbf{o}_1, \mathbf{o}_2, \cdots, \mathbf{o}_n)
\end{align}

\subsection{Implementation Details}
\begin{figure*}[tb]
  \centering
  \includegraphics[width=1.0\textwidth]{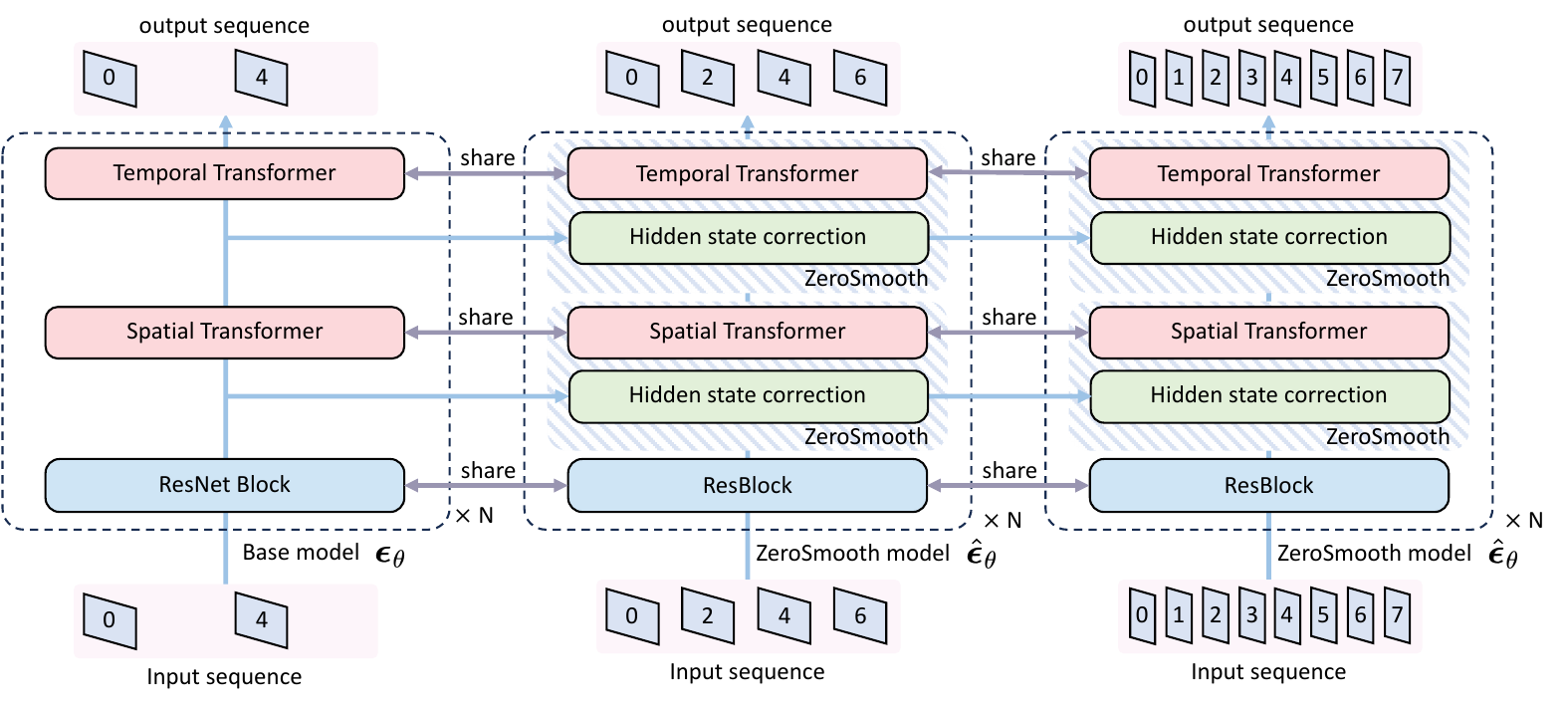}
  \caption{
    An illustration of three stages self-cascaded model.
  }
  \label{fig:cascade-4x}
\end{figure*}
We describe the implementation details that is not included in the main text in this subsection, showcasing the adaptation methods for all the used models including VideoCrafter2, LaVie and StableVideoDiffusion. 

\paragraph{Modules to apply ZeroSmooth} Modern video diffusion UNet includes 6 categories of neural network modules. 
% spatial ResNet block, temporal ResNet block, spatial self-attention, spatial cross-attention, temporal self-attention and temporal cross-attention. 
ZeroSmooth is applied to spatial self-attention, spatial cross-attention, temporal self-attention and temporal cross-attention. The remaining modules including spatial ResNet block and temporal ResNet block are kept the same to the base generator. We use our method to all the modules in the selected categories for all denoising timesteps. When computing the key and the value, cross-attention modules uses text prompts or image embedding in text-to-video and image-to-video respectively. Since the condition is repeated for all frames, our hidden state correction will not change the value of these conditions. 

\paragraph{Stochasticity control} We use a noise scheduling strategy to reduce the stochasticity during denoising sampling. Consider a $N$-stage self-cascaded ZeroSmooth model that includes $\{\boldsymbol{\epsilon}^1_\theta(\cdot): \mathbb{R}^{t_0 \times hwc} \rightarrow \mathbb{R}^{t_0 \times hwc}, \boldsymbol{\epsilon}^2_\theta(\cdot): \mathbb{R}^{nt_0 \times hwc} \rightarrow \mathbb{R}^{nt_0 \times hwc}, \cdots, \boldsymbol{\epsilon}^N_\theta(\cdot): \mathbb{R}^{n^{N-1}t_0 \times hwc} \rightarrow \mathbb{R}^{n^{N-1}t_0 \times hwc}\}$. $n$ is the frame interpolation scale. Let $\epsilon^N \in \mathbb{R}^{n^{N-1}t_0 \times hwc}$ denote the additional standard Gaussian noise introduced to the denoising iteration of $\boldsymbol{\epsilon}_\theta^N(\cdot)$ for a specific denoising timestep. Let the lower script $i$ denote the noise in $i$-th frame. Then, the additional noised used for $\boldsymbol{\epsilon}_\theta^s(\cdot)$ is sampled from $\epsilon^N$ via
\begin{equation}
    \epsilon_i^s = \epsilon_{in^{N-s}}^N.
\end{equation} This strategy ensures the noise at the corresponding frame to be the same in all self-cascaded stages. 

\paragraph{VideoCrafter2} VideoCrafter2 uses an additional frame-per-second condition to control the motion speed presents in the video. Given a fps condition for the base generator. We multiply it with the interpolation scale to get the fps condition of a spefic self-cascaded stage.

\paragraph{StableVideoDiffusion} StableVideoDiffusion uses two additional conditions including fps and motion bucket id. Motion bucket id controls the the motion amplitude in the video. Since the overall motion is kept unchanged in high frame video, the motion bucket id is changed in all stages. We apply a similar strategy to set fps like in VideoCrafter2. 

Unlike text-to-image video diffusers, StableVideoDiffusion concatenate a noisy reference image to every frame to provide a guidance for visual appearance. The noise strength added to the reference image is callded augmentation strength. In our method, we use a uniform augmentation strength for all stages. StableVideoDiffusion uses classifier-free guidance \cite{classifier-free} during sampling. The guidance scale is linear monotonically decreasing along the frame number. To adapt this feature in ZeroSmooth, we fix the highest and the lowest guidance scale and use linear interpolation to get the guidance scale for the high frame rate video.

\paragraph{Self-cascaded model}

Our experiment includes 2$\times$ and 4$\times$ higher frame rate generation. We use two stages for the 2$\times$ experiment and use  three stages for the $4\times$ experiment. Every stage will generate $2\times$ higher frame rate videos. A illustration of the three stages model is shown in Fig.\ref{fig:cascade-4x}. We pass the corrected states of the second stage $\hat{\mathbf{h}}, \hat{\mathbf{Q}}, \hat{\mathbf{K}}, \hat{\mathbf{V}}$ to calibrate the hidden states in the third stage.

\paragraph{Controlling correction strength}
The weight controlling correction strength is defined as
\begin{equation}
    w_t = 0.8\sqrt{t / T},
\end{equation}
where $T$ is the maximum denoising timestep. We find in the early stage of denoising, the model is more tolerant to hidden states correction. Therefore we build a monotonically decreasing $w_t$. In the late stage of denoising, the strength of hidden states correction becomes small. 
% Note that StableVideoDiffusion adopts euler discrete sampler, 

\paragraph{Correcting color tone} An additional normalization is applied to avoid abrupt color tone change in the interpolated frames. We find this helps in some special cases. Specifically, after each denoising iteration. We use adaptive instance normalization to calibrate the statistics of frames in the high frame rate video to match the corresponding key frames. 

\subsection{Experiment Settings}
% \paragraph{Hyperparameters}
\begin{table*}[tb]
    \centering
        \begin{tabular}{lccc}
        \toprule
            Hyperparameter & \textbf{VideoCrafter2} & \textbf{LaVie} & \textbf{StableVideoDiffusion} \\
            \midrule
            Frame resolution    & $320\times512$ & $320\times512$ & $576\times1024$ \\
            Frame number        & (16, 32, 64) & (16, 32, 64) & (14, 28, 56) \\
            FPS condition       & 8 & - & 7 \\
            Motion bucket id    & - & - & 180 \\
            Aug. strength       & - & - & 0.02 \\
            Inference steps     & 50 & 50 & 25 \\
            Eta                 & 1.0 & 1.0 & - \\
            Temp. attn. overlap & 8 & 8 & 7 \\
            Sampler             & DDIM samplre & DDIM sampler & Euler discrete sampler \\
            Guidance scale      & 12.0 & 7.5 & 7.5 \\
        \bottomrule
        \end{tabular}
        \caption{Hyperparameters of ZeroSmooth.}
        \label{tab:hyperparameters}
\end{table*}

The hyperparameter used for the experiment is shown in Tab.\ref{tab:hyperparameters}. Aug. strength denotes noise augmentation strength used for StableVideoDiffusion in image-to-video. FPS condition is an input for VideoCrafter2 and StableVideoDiffusion. Temporal attention overlap determines how many frames are shared between attention windows in temporal transformers. Eta controls the noise added for every denoising step in DDIM sampler. We use a stochastic samping trajectory for StableVideoDiffusion, no additional randomness is added after the initialization during sampling.

When using LDMVFI, we apply DDIM sampler. The inference timestep to 50 and eta is set to 0. Classifier-free guidance scale is 0 for LDMVFI. The model achieves $2\times$ frame interpolation. When using LaVie interpolation model, we also use DDIM sampler. The inference timestep is 25 and eta is 0. Classifier-free guidance scale for LaVie interpolator is 4.

We deploy all the models in fp16 precision during inference. 32 NVIDIA V100 32GB GPUs are used to sample the results. Due to the memory limit, we use flash attention to reduce the VRAM requirements. The transformer hidden states in the previous stage are offloaded to CPU memory and is thrown to VRAM when it is need.

% \paragraph{Inference details}
\subsection{Additional Visual Comparisons}

More visual comparisons between tuning-free methods and comparions between ZeroSmooth and training-based methods are shown below.

% \begin{figure*}[tb]
%   \centering
%   \includegraphics[width=1.00\textwidth]{res/compare-tuning-free.pdf}
%   \caption{
%     Qualitative comparison bettween our method, Gen-2 \cite{gen-1}, VideoCrafter1 \cite{chen2023videocrafter1} and ModelScope \cite{modelscope}.
%    }
%   \label{fig:compare-funing-free}
% \end{figure*}

\begin{figure}[htbp]
    \makebox[\textwidth][c]{
    \includegraphics[width=1.4\textwidth]{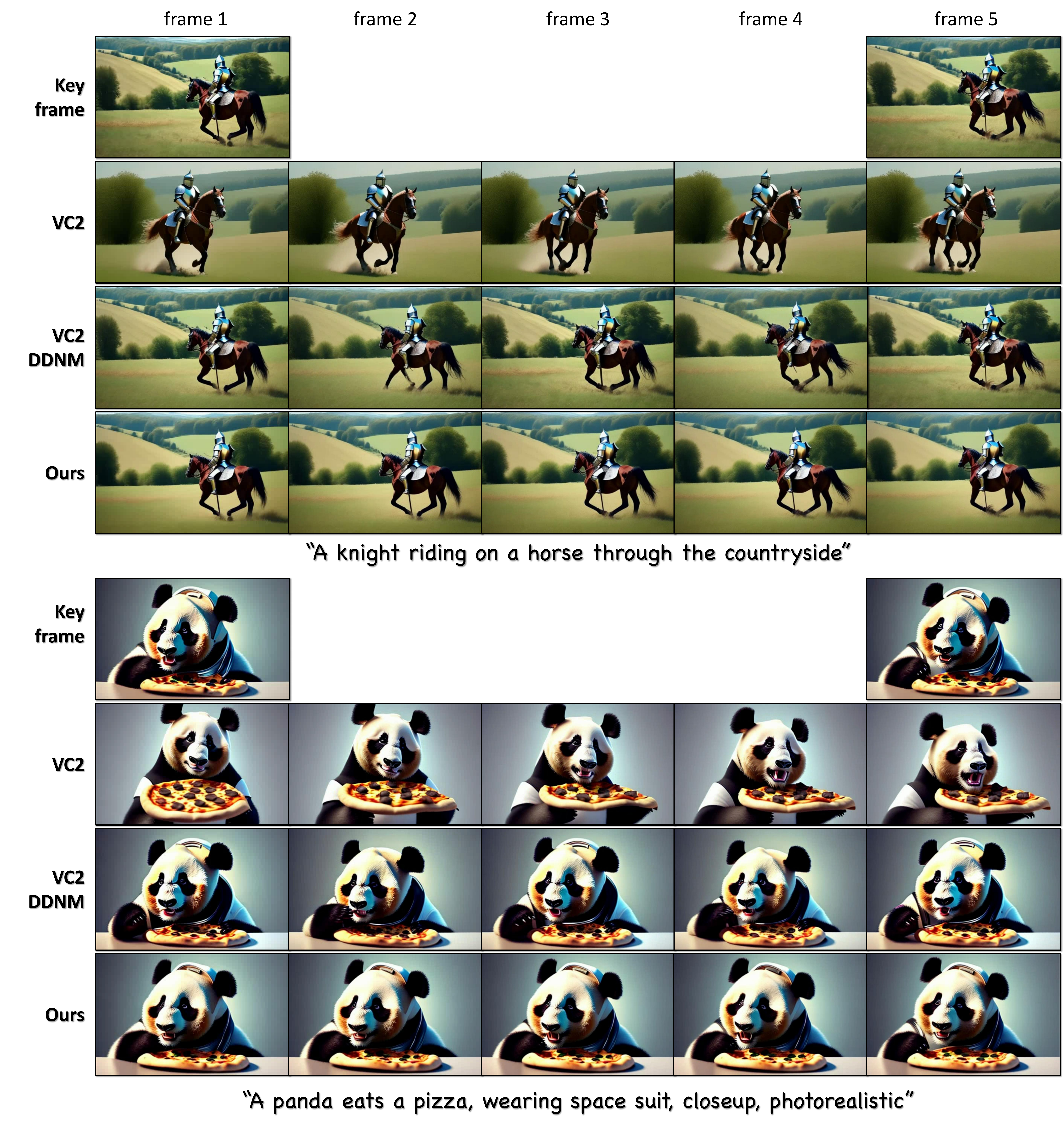}}  
    \caption{Visual comparison between our method and other tuning-free baselines on VideoCrafter2.} 
    \label{fig:appendix_fig_1}
\end{figure}

% \begin{figure*}[tb]
%   \centering
%   \includegraphics[width=1.00\textwidth]{res/compare-tuning-free.pdf}
%   \caption{
%     Qualitative comparison bettween our method, Gen-2 \cite{gen-1}, VideoCrafter1 \cite{chen2023videocrafter1} and ModelScope \cite{modelscope}.
%    }
%   \label{fig:compare-funing-free}
% \end{figure*}

\begin{figure}[htbp]
    \makebox[\textwidth][c]{
    \includegraphics[width=1.4\textwidth]{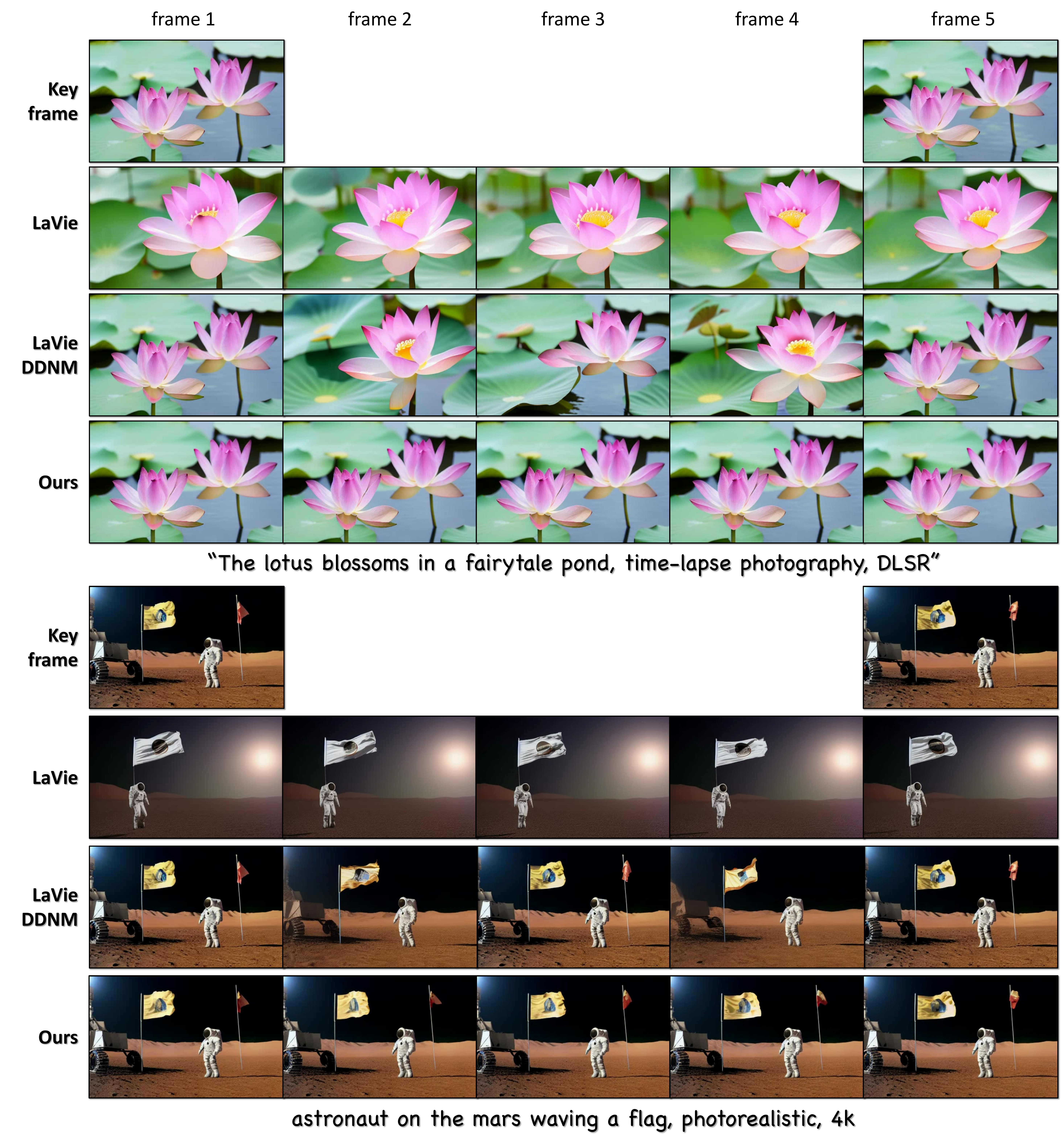}}  
    \caption{Visual comparison between our method and other tuning-free baselines on LaVie\cite{wang2023lavie}.} 
    \label{fig:appendix_fig_1}
\end{figure}

% \begin{figure*}[tb]
%   \centering
%   \includegraphics[width=1.00\textwidth]{res/compare-tuning-free.pdf}
%   \caption{
%     Qualitative comparison bettween our method, Gen-2 \cite{gen-1}, VideoCrafter1 \cite{chen2023videocrafter1} and ModelScope \cite{modelscope}.
%    }
%   \label{fig:compare-funing-free}
% \end{figure*}

\begin{figure}[htbp]
    \makebox[\textwidth][c]{
    \includegraphics[width=1.4\textwidth]{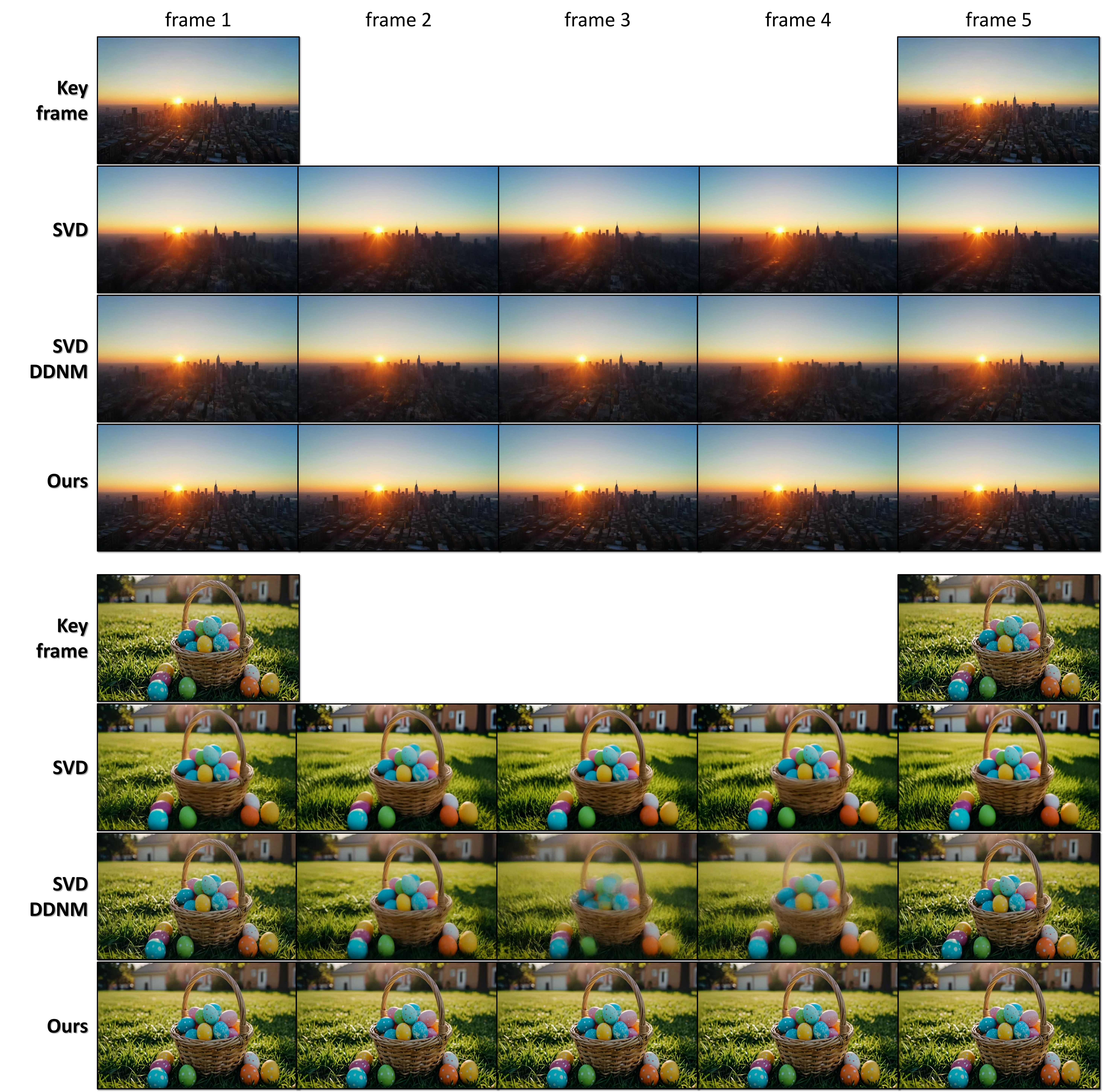}}  
    \caption{Visual comparison between our method and other tuning-free baselines on StableVideoDiffusion.} 
    \label{fig:appendix_fig_1}
\end{figure}

% \begin{figure*}[tb]
%   \centering
%   \includegraphics[width=1.00\textwidth]{res/compare-tuning-free.pdf}
%   \caption{
%     Qualitative comparison bettween our method, Gen-2 \cite{gen-1}, VideoCrafter1 \cite{chen2023videocrafter1} and ModelScope \cite{modelscope}.
%    }
%   \label{fig:compare-funing-free}
% \end{figure*}

\begin{figure}[htbp]
    \makebox[\textwidth][c]{
    \includegraphics[width=1.4\textwidth]{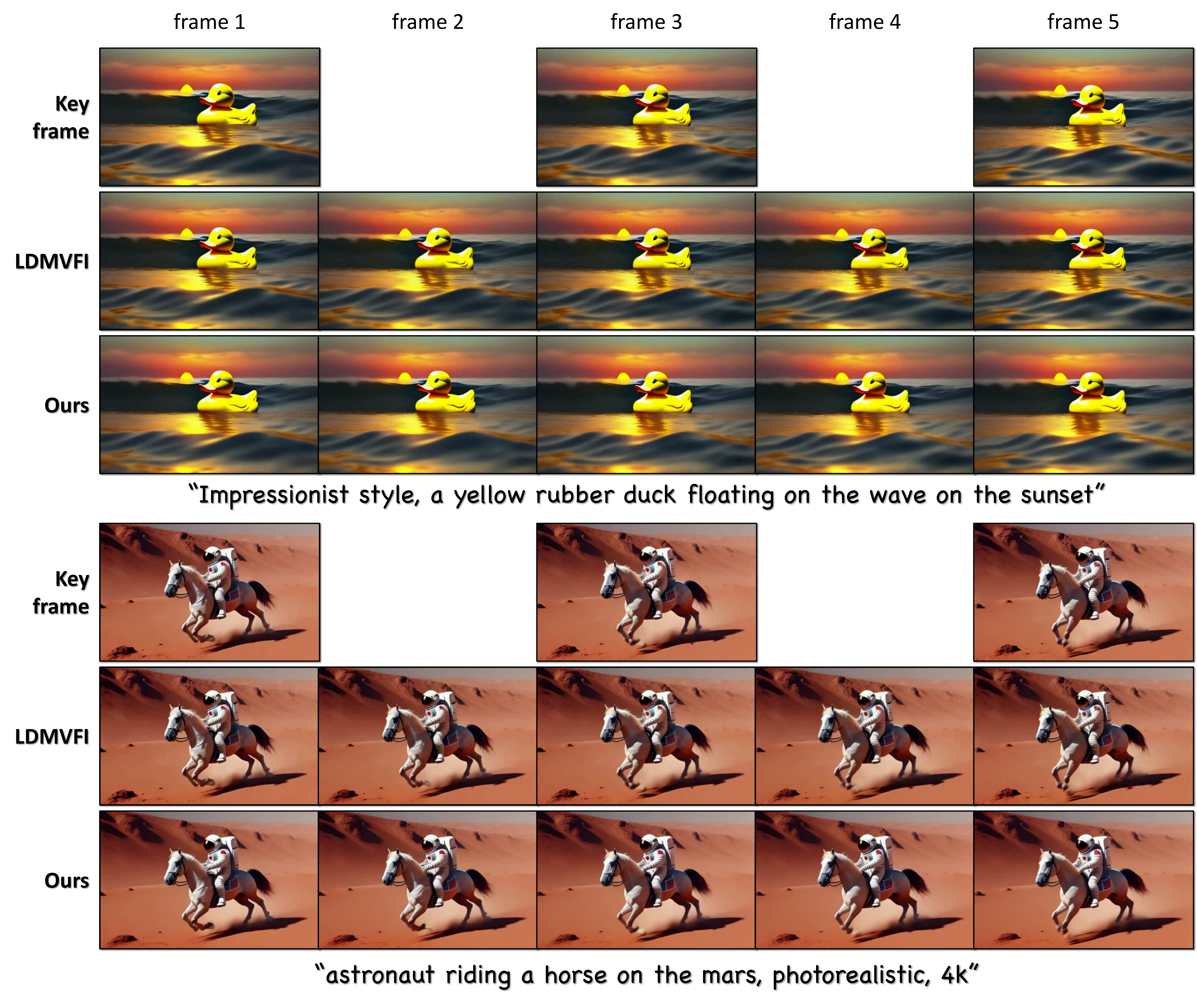}}  
    \caption{Visual comparison between our method and LDMVFI \cite{danier2023ldmvfi}.} 
    \label{fig:appendix_fig_1}
\end{figure}

% \begin{figure*}[tb]
%   \centering
%   \includegraphics[width=1.00\textwidth]{res/compare-tuning-free.pdf}
%   \caption{
%     Qualitative comparison bettween our method, Gen-2 \cite{gen-1}, VideoCrafter1 \cite{chen2023videocrafter1} and ModelScope \cite{modelscope}.
%    }
%   \label{fig:compare-funing-free}
% \end{figure*}

\begin{figure}[htbp]
    \makebox[\textwidth][c]{
    \includegraphics[width=1.4\textwidth]{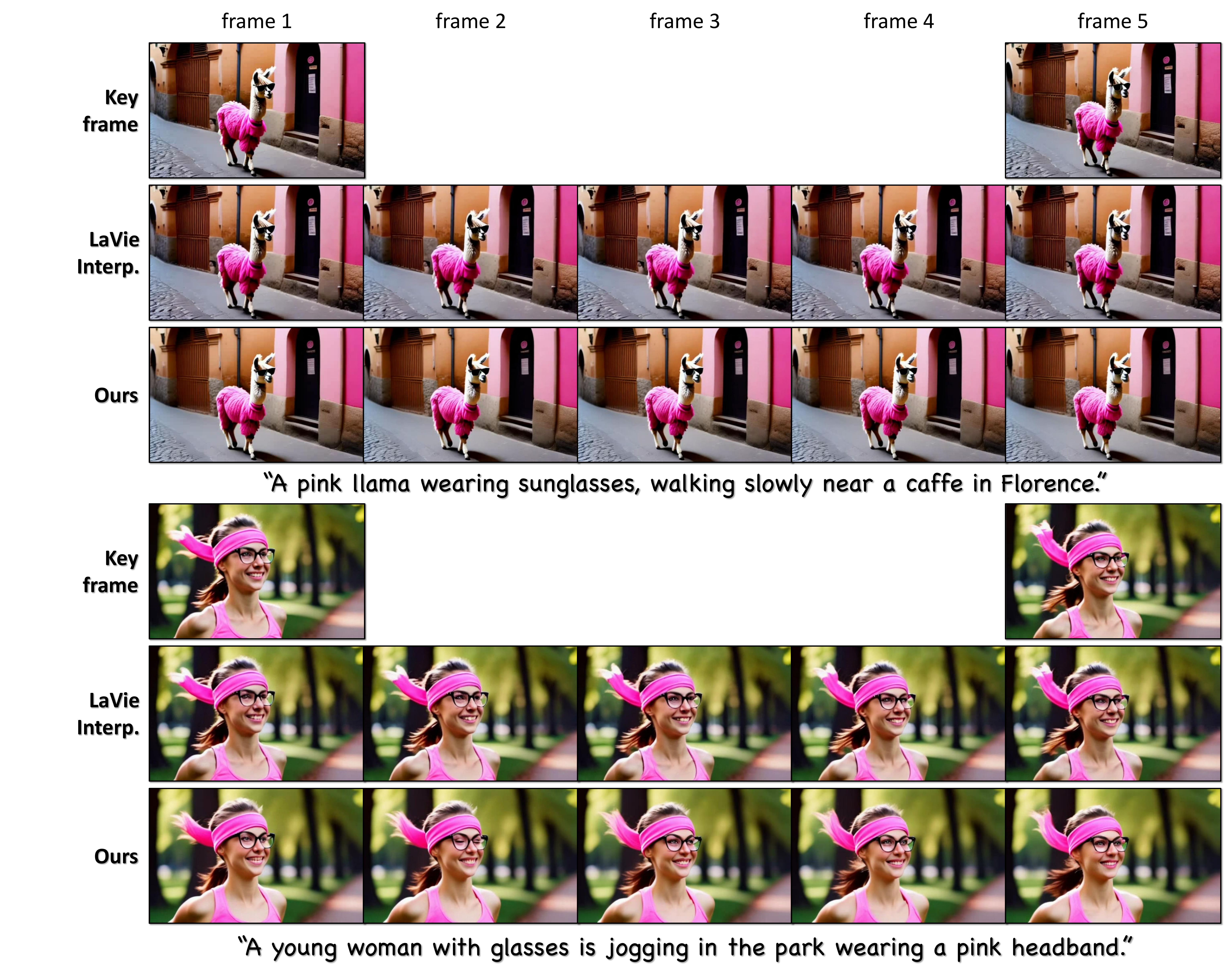}}  
    \caption{Visual comparison between our method and LaVie interpolation model \cite{wang2023lavie}.} 
    \label{fig:appendix_fig_1}
\end{figure}

% \newpage
% 
% Optionally include supplemental material (complete proofs, additional experiments and plots) in appendix.
% All such materials \textbf{SHOULD be included in the main submission.}

%%%%%%%%%%%%%%%%%%%%%%%%%%%%%%%%%%%%%%%%%%%%%%%%%%%%%%%%%%%%

% \input{sec/8_checklist}

\end{document}